%% file: main.tex
\documentclass[letterpaper, 10 pt, conference]{ieeeconf}  

\IEEEoverridecommandlockouts                              

\usepackage{graphics} 
\usepackage{epsfig} 
\usepackage{float}
\usepackage{svg}

\usepackage[caption=false,font=footnotesize]{subfig}

\usepackage{amsmath} 
\usepackage{amssymb}  

\usepackage{graphicx}               

\usepackage[ruled]{algorithm2e}

\usepackage[bookmarks=false, hidelinks]{hyperref}
\usepackage[capitalise,nameinlink]{cleveref}

\usepackage{siunitx}

\usepackage{xspace}
\newcommand{\eg}{\hbox{\emph{e.g.}}\xspace}
\newcommand{\ie}{\hbox{\emph{i.e.}}\xspace}
\newcommand{\etc}{\hbox{\emph{etc}}\xspace}

\usepackage{xcolor}
\usepackage[export]{adjustbox}

\newcommand{\theo}[1]{{\color{black}#1}}
\newcommand{\micha}[1]{{\color{black}#1}}

\title{\LARGE \bf
Predictive and Robust Robot Assistance for Sequential Manipulation
}

\author{Theodoros Stouraitis$^{1}$ and Michael Gienger$^{1}$
\thanks{$^{1}$Theodoros Stouraitis and Michael Gienger are with Honda Research Institute Europe (HRI-EU), 63073 Offenbach am Main, Germany. {\tt\small stoutheo@gmail.com, \tt\small michael.gienger@honda-ri.de}}%
\thanks{
The authors would like to thank Stephan Hasler for his help with the experimental setup.}
}

\begin{document}

\maketitle
\thispagestyle{empty}
\pagestyle{empty}

\begin{abstract}
This paper presents a novel concept to \micha{support physically impaired} humans in daily object manipulation tasks with a robot. 
Given a user's manipulation sequence, we propose a predictive model that uniquely casts the user's sequential behavior as well as a robot support intervention into a hierarchical multi-objective optimization problem. A major contribution is the prediction formulation, which allows to \theo{consider} several different future paths concurrently. The second contribution is the encoding of a general notion of constancy constraints, which allows to consider dependencies between \theo{consecutive or far apart} keyframes \theo{(in time or space)} of a sequential task.
We perform numerical studies, simulations and robot experiments to analyse and evaluate the proposed method  in several table top tasks where a robot supports impaired users by predicting their posture and proactively re-arranging objects.

\end{abstract}

\section{Introduction}
\label{sec:intro}

\input{sections/intro}

\section{Related work}
\label{sec:related_work}
\input{sections/related_work}

\section{Problem Statement}
\label{sec:prob_state}
\input{sections/Problem_statement}

\section{Method}
\label{sec:methods}
\input{sections/Method}

\section{Evaluations and Experiments}
\label{sec:results}

\input{sections/Evaluations}


\section{Summary and discussion}
\label{sec:conclusion}
\input{sections/conclusion}



\bibliographystyle{IEEEtran}
\bibliography{references.bib}

\end{document}

%% file: sections/intro.tex
Assistive robots can play a crucial role for more than 200 million humans that need assistance with activities of daily living (ADLs)~\cite{Who2011}, such as getting out of bed, going to the restroom, preparing a meal, \etc. Towards addressing this societal challenge~\cite{sendhoff2020cooperative}, the areas of physical Human-Robot-Interaction (pHRI)~\cite{ajoudani2018progress} and rehabilitation robotics~\cite{yan2015review} have developed approaches for 
ergonomic~\cite{kim2017anticipatory}, safe~\cite{mainprice2013human}, 
and personalised~\cite{Gordon2022} support. Key challenges studied are \theo{capturing human motion~\cite{fang2020wearable}}, anticipating human behaviour~\cite{koppula2016anticipatory}, modelling  robot behaviour influence to the human behaviour~\cite{trautman2010unfreezing}, 
and coping with variations~\cite{erickson2018deep, li2022set} of human behavior due to factors like disabilities, partial observability, \etc.

\begin{figure}
    \centering
    \includegraphics[width=0.85\linewidth]{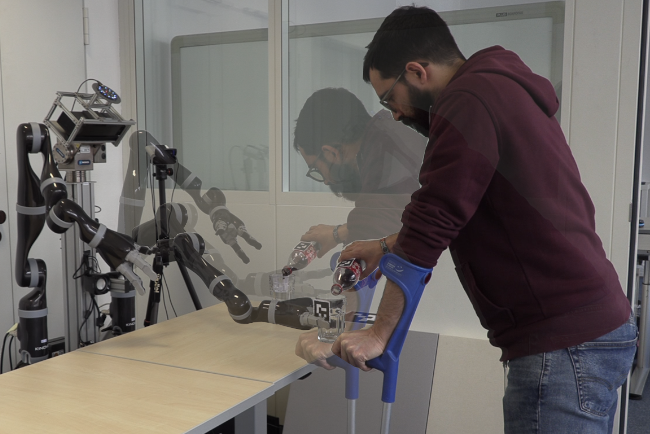}
    \vspace{-2.5mm}
    \caption{Illustration of the presented concept in a supported pour-bottle-into-glass task. A robot 
    intervenes to move the glass (transparent view) to a location on the table that enables an impaired user to comfortably pour the content of a bottle into it.
    The supported human (solid view) is enabled to perform the task optimally given an ergonomic metric. In the non-supported case, the (transparent) human needs to get into an uncomfortable posture. } 
    \label{fig:teaser}
    \vspace{-6mm}
\end{figure}

\begin{figure*}[t]
    \vspace{1mm}
    \centering
    \includegraphics[width=0.9\linewidth]{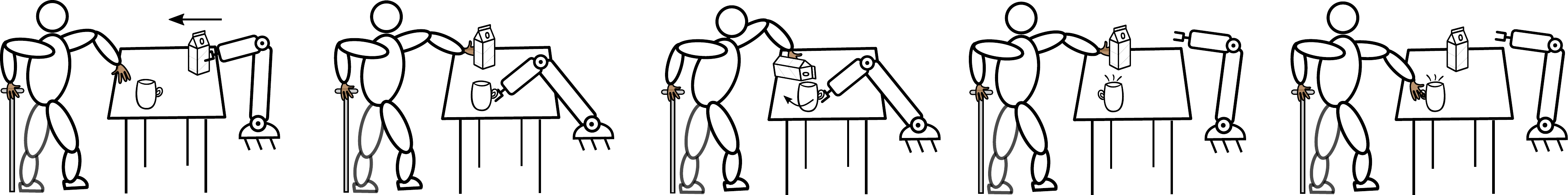}
    \caption{Illustration of a sequential manipulation where the robot assists a human to prepare a beverage by pushing the milk and rotating the mug.}
    \label{fig:concept_seq_manip}
    \vspace{-6mm}
\end{figure*}

In this paper, we study these challenges in the context of manipulation. Our focus is on temporally extended ADLs that require sequential manipulation capabilities.  
As a typical example, consider an impaired human (\micha{immobilized left arm}) preparing a beverage, as shown in~\cref{fig:teaser} and~\cref{fig:concept_seq_manip}. 
\micha{This task requires bimanual operations, 
and due to the injury the human has rely on the robot support.
To provide optimal support}, the robot needs to predict the human's actions (discrete decisions) and their outcome (continuous states)~\cite{busch2018planning}, assess the possible human postures~\cite{van2020predicting}, and intervene to assist. Key to this scenario is that the prediction needs to consider a set of possible futures, as accurate predictions of the human's actions might not be possible. 

\micha{In this work, we assume that the action schedule (task-level coordination~\cite{gombolay2015decision}) is known, and we focus on interventions that (i) adjust continuous geometric quantities, \eg  change the pose of the mug, (ii) parallel actions, \eg while the human picks the bottle the robot fetches the mug.  }
Following the "assist-as-needed" principle\footnote{\vspace{-0.5mm}"assist-as-needed" indicates that assistance should be minimal to motivate the voluntary contribution of the impaired human (patient).},
we view the robot interventions as preparatory actions~\cite{chang2008preparatory} that enable or ease subsequent actions and motions of the impaired human. This translates into a method that has as backbone a physics-based prediction mechanism, which allows us to investigate the effects of possible (robot) interventions. 
The prediction mechanism can be considered as part of combined task and motion planning (TAMP)~\cite{ garrett2021integrated}, where the task schedule (linear or tree sequence of modes in the form of constraints) is given. Adapting online the sequences of modes~\cite{toussaint2022sequence, StouraitisTRO2020} is beyond the scope of this work. 
Thus, we ask ourselves the question "\textit{how to turn \micha{an action schedule} into a prediction of the human's plan while considering; (a) set of possible futures, and (b) the influence of intervention by a robot?}

\theo{
We focus on the low-level geometric choices in keyframes, rather than computing trajectories as in Trajectory Optimization methods. 
\micha{Keyframes represent important transitions within the action sequence such as contact transitions or the start of a new task-level trajectory. They describe the subgoal structure of the task.}
Based on keyframes we developed} 
an efficient constraint resolution method that; (i) can consider both a linear and a tree-like sequence of modes, (ii)  holds robot actions and their effects, (iii) scales to large sequences of high dimensional systems, like the DoFs of the human body, and (iv) is reactive in the sense that predictions can be updated within few hundreds of \micha{milliseconds}.
\micha{We call this,} reactive Tree-Consistent-Constraint (TCC) resolution method. 
The \micha{action} schedule defines a set of constraints,
and TCC predicts a sequence (linear or tree) of keyframes that satisfy them. The interdependecies between these keyframes are formed with spatio-temporal constraints that create links between bodies in task space, such as objects, end-effectors, \etc. Interventions are represented with open variables that are optimised to satisfy secondary objectives, such as postural preferences.
The contributions of our work are:
\begin{itemize}
    \item \micha{The} reactive Tree-Consistent-Constraint (TCC) method that can simultaneously \micha{compute} a set of possible futures based on \theo{temporal and concurrent dependencies.}
    \item A computation formalism based on 
    Damped Least Squares (DLS) to cast cooperative sequential manipulation as a single underactuated system where robot assistance (\ie interventions) lies in the null-space of manipulation sequences. 
    \item Demonstrating the capabilities of \micha{TCC} to predict \micha{and optimize} keyframes of uncertain human manipulation sequences \micha{including assistive} robot interventions. 
\end{itemize}

\noindent In~\cref{sec:related_work}, we review the related work, in~\cref{sec:prob_state} we formally set the problem and in~\cref{sec:methods}, we present our method. In~\cref{sec:results} we evaluate the proposed method and demonstrate its capabilities in real world setups.

%% file: sections/related_work.tex
\subsection{Human-aware \micha{prediction,} planning and control}

An extensively studied prediction area is ground-level 2D trajectories~\cite{rudenko2020human}. In contrast to that, we focus on predicting articulated full body motions in  manipulation sequences. 
Prior model-based approaches demonstrate local adaptation of the robot posture that accounts for the overloading human joint torques~\cite{kim2017anticipatory, peternel2017towards}. 
Others learned models to predict human motion and to adapt the robot plans. In~\cite{campbell2019probabilistic} human-robot interaction primitives are used to select and coordinate the robot movement with the recognized human actions, while in~\cite{mainprice2013human, jain2020anticipatory} safe (in terms of collision)  concurrent (human and robot) reaching motions are demonstrated.
These works studied prediction of articulated bodies, yet they only consider short-term interactions without sequential dependencies, \theo{\ie interdependencies in a series of actions.}

Most relevant to our work are cooperative TAMP methods~\cite{busch2018planning, kratzer2020prediction} that plan decisions and motions of both the human and the robot (articulated bodies) towards finding offline the optimal robot manipulation plans. In our approach predictions can be updated on-the-fly, as in~\cite{kratzer2020prediction}, yet ours also enables simultaneous consideration of a set of possible future sequences, rather than a single linear sequence.

\subsection{Spatio-temporal constraints}

Prior work has proposed various constraint formulations to efficiently synthesize whole-body  motion plans. These motions exhibit complicated dependencies across concurrent tasks (same time-slice)~\cite{berenson2009pose}, and along the horizon (between keyframes)~\cite{toussaint2022sequence}. This led to the question: "\textit{How to model concurrent and sequential dependencies with constraints?}"

Explicit geometric constraints (distances, friction cone, \etc) have been used in ~\cite{bouyarmane2010static} to form and solve a multi-robot co-manipulation problem. 
Task-space-region-chains~\cite{berenson2009pose} were proposed to link together a set of workspace-goal-regions (similar to task-intervals~\cite{gienger2006exploiting}) and form a manifold where RRT-based planners can search for whole-body manipulation paths. Both approaches can consider concurrent dependencies, \eg closed kinematics chains, multiple robots postures, \etc. 
Yet, they did not consider sequential representations of several connected postures.

Temporal dependencies---a crucial modelling aspect of  sequential motions---were \theo{analytically} modelled by linking consecutive footsteps 
given symmetric gait patterns in~\cite{kanoun2011planning}. 
The Contact-Consistent Elastic Strips framework~\cite{chung2015contact} combined the concept of task-intervals~\cite{gienger2006exploiting} or task-space-region~\cite{berenson2009pose} with contact-consistent constraints that couple the consecutive robot poses to generate multi-contact locomotion plans. More generally temporal dependencies can be encoded as \textit{constancy constraints}~\cite{toussaint2022sequence} either by pair-wise coupling constraints or by sharing of DoFs across time-slices~\cite{toussaint2021co}. We use task-intervals~\cite{gienger2006exploiting} and we broaden the notion of \textit{constancy constraints} to capture both temporal~\cite{toussaint2022sequence} and concurrent dependencies. This enables us to optimize manipulation planning trees, rather than linear sequences.

%% file: sections/Problem_statement.tex
Our goal is to investigate the influence of robot assistive actions in sequential manipulation tasks that include a set of objects and are performed by an impaired human. To do that, we break-down the problem into two \theo{focal points}.

Following the multi-modal motion planning (MMMP) notation, we denote the configuration space (C-space) of $K$ movable objects\footnote{\vspace{-1mm}\label{note1}The superscripts on $O$s, $\theta$s and $B$s are indexes and not exponents.} as $\mathcal{O} = O^1 \times ... \times O^K \subseteq SE(3)^K$, the C-space of the human is $H \subseteq \mathbb{R}^\alpha$. Hence, the configuration space of the prediction problem is $ \mathcal{P} = H \times \mathcal{O} \subseteq SE(3)^K \times \mathbb{R}^{\alpha}$, in which we are searching for paths (to be executed by a human) that have a known initial configuration $\mathbf{p}_0 \in \mathcal{P}$ and comply with constraints of the sequential manipulation task. These consider task-scenario specifications---\eg impairment, connected objects, stable contacts, constant grasps, \etc---and temporal transitions---\eg whether an object can move or not, whether an object can slide in a hand or not, \etc---such that the path is feasible. The former partitions the configuration space, while the latter shapes how the partitions of the configuration space are linked together. We represent specifications and transitions with constraints as $\mathbf{T}_{\mathcal{P}}(\cdot) \leq 0 $, where $\mathbf{T}_{\mathcal{P}}(\cdot) : \mathcal{P}^\gamma \xrightarrow{} \mathbb{R}^\gamma$ and $\gamma$ is the total number of constraints (more details are provided in~\cref{sec:methods}).

\subsubsection*{\underline{Many possible futures}}
A sequential manipulation task is conditional on the preferences of the human that performs it. For example, a human might prefer to first pick up the milk and then the mug (see~\cref{fig:concept_seq_manip}) or vice versa, or a human might prefer to power-grasp the mug, rather than picking it up from the handle using a pinch-grasp. We describe these preferences by the latent variable $\boldsymbol{\theta}$, which is one out of a finite set of many predefined preferences\textsuperscript{\ref{note1}} $\Theta = \{\boldsymbol{\theta}^1, \boldsymbol{\theta}^2,...\boldsymbol{\theta}^\iota\}$, also referred to as modes in MMMP. 
\theo{The human preferences are not likely to be known in advance, hence a prediction of $N$ future manipulation steps should be able to consider many preferences concurrently.} 

\textbf{First focal point:}
Develop a prediction method that \theo{\textit{predicts a set} $\mathcal{S}$ of feasible future paths in $\mathcal{P}$. This set is}
\begin{equation}
  \label{eq:pred-only}     
  \mathcal{S} = \left\{
  \mathbf{h}_{1:N}, \mathbf{o}^k_{1:N}:  \mathbf{T}_\mathcal{P}(\mathbf{p}_0, \mathbf{h}_{1:N}, \mathbf{o}^k_{1:N}; \Theta) \leq 0 
  \right\},
\end{equation}
where $\mathbf{h}_{1:N}$ and $\mathbf{o}^k_{1:N}$ are linear sequences of configurations for the human and the movable objects, respectively.

\subsubsection*{\underline{Influence of interventions}}
\label{par:influence_inter}
Let us consider that an assistive agent (robot)---with C-space $Q \subseteq \mathbb{R}^\beta$---can also manipulate the same movable objects. Including  the robot agent results in an extension of the configuration space to
$ \mathcal{X} = \mathcal{P} \times Q \subseteq SE(3)^K \times \mathbb{R}^{\alpha + \beta}$, with total number of dimensions represented by $d_\mathbf{x} \in \mathbb{R}$. Thus, we extend $\mathbf{T}_{\mathcal{P}}(\cdot)$ to  $ \mathbf{T}_\mathcal{X}(\cdot) \leq 0 $ with $\mathbf{T}_\mathcal{X}(\cdot) : \mathcal{X}^\gamma \xrightarrow{} \mathbb{R}^\gamma$, such that it can also encode robot specific constraints. Using the above we can now define a \textit{prediction set} $\mathcal{S}'$ of feasible future paths that are influenced by the robot's interventions, as
\begin{equation}
  \label{eq:pred-intervene}  
  \hspace{-1.75mm}
  \mathcal{S}' = \left\{
  \mathbf{h}_{1:N}, \mathbf{o}^k_{1:N}:  \mathbf{T}_\mathcal{X}(\mathbf{x}_0, \mathbf{h}_{1:N}, \mathbf{o}^k_{1:N}, \mathbf{q}_{1:N}; \Theta) \leq 0 
  \right\},
\end{equation}
where $\mathbf{x}_0 \in \mathcal{X}$ is a known initial configuration, and our goal is to optimize the robot's interventions such that $\mathcal{S}'$ has lower cost than $\mathcal{S}$ according to a metric of the human state. 
 
\textbf{Second focal point:}
Find a single linear sequence of robot configurations $\mathbf{q}_{1:N}$ that minimizes a metric for all possible future paths of the human. This can be expressed as 
\begin{IEEEeqnarray}{CCCCC}
    \IEEEyesnumber\label{eq:RO} 
    \IEEEyessubnumber \label{eq:RO02}
    \min_{\mathbf{q}_{1:N}}  & ~\mathbf{c}\left(\mathbf{h}_{1:N}\right)~\hfill \\
    ~\text{s.t.}~  
    \IEEEyessubnumber \label{eq:RO03} 
    & \forall~\mathbf{h}_{1:N} \subset \mathcal{S}' \hfill
\end{IEEEeqnarray}
where $\mathcal{S}'$ contains all the possible future paths of the human which according to~\eqref{eq:pred-intervene} depends on the robot plan $\mathbf{q}_{1:N}$, \ie robot interventions. Cost function $\mathbf{c}(\cdot)$ \theo{is} a metric of the human state, \eg comfort, ergonomics, effort, \etc. Note that~\eqref{eq:RO} is a Robust Optimization (RO) problem as we seek to find a solution that minimizes a metric for all possible futures. Specifically, this is a \textit{multiforecast model predictive optimization} formulation (see sec. 9.9.3 in~\cite{kochenderfer2022algorithms}).

%% file: sections/Method.tex
\begin{figure}[t]
    \centering
    \def\svgwidth{0.95\linewidth}\input{figures/methods/concept-seqManip-drawing} 
    \vspace{-2mm}
    \caption{\theo{Illustration of the constraints used to describe the first three keyframes of~\cref{fig:concept_seq_manip} with  the robot and its actions being omitted for clarity. \textit{Task  constraints} are used only within keyframes and \textit{constancy constraints} are used between keyframes, hence they are illustrated with arrows to link timeframes and diamonts to indicate which dimensions remain constant.}}
    \label{fig:constraints_concept}
    \vspace{-5mm}
\end{figure}
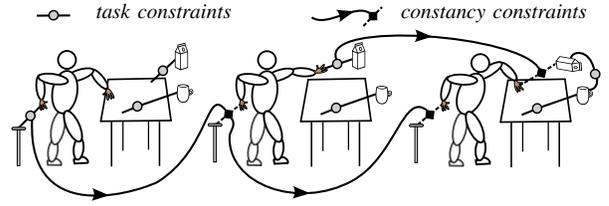

To achieve the two \theo{focal points} stated above, we propose a model-based optimisation approach that has the two following key aspects. First, the constraints $ \mathbf{T}_\mathcal{X}(\cdot) \leq 0 $ are \textit{task constraints} for each keyframe and \textit{constancy constraints} that encapsulate a notion of invariance between keyframes. \theo{\cref{fig:constraints_concept} shows these types of constraints being used to describe a sequence of manipulation actions.} Second, we developed a barrier method that uses a Levenberg–Marquardt update to resolve all the constraints concurrently (even across parallel future paths). \theo{Solve an optimization problem with all constraints shown in \cref{fig:constraints_concept} results in the corresponding keyframes of~\cref{fig:concept_seq_manip} (three first).} This allows us to regard all possible futures (via the constraints) as primary \theo{objectives} and treat the robot's assistance as secondary \theo{objectives} and hence, find a single linear sequence of robot interventions that improves the human state for all possible futures. Next, we describe these two aspects in detail.

\subsection{Spatio-temporal constraints}
\label{subsec:constraints}

To ease the description that follows, we introduce $L$ number of bodies that belong to the kinematic chains of the human or the robot (functions with domain $H$ and $Q$)\footnote{Note, that $Bs$ are redundant, they are not represented explicitly in our method and are only used here to ease the description of constraints.}, and we denote their C-space\textsuperscript{\ref{note1}} as $B^1, ..., B^L \subseteq SE(3)$.

\subsubsection*{\underline{Task constraints}} We use \textit{task constraints} to \theo{analytically} model each manipulation step (time-slice).
These can be written as $\mathbf{g} \left(\mathbf{x}_i\right) \leq 0$, where $\mathbf{g}(\cdot):\mathcal{X}\xrightarrow{}\mathbb{R}^{\gamma_g} $ with $\gamma_g \in \mathbb{R}$ and $i$ indicates the current keyframe in a sequence.
The explicit form, in terms of input arguments, of $\mathbf{g}(\cdot)$ is   
\begin{equation}
    \label{eq:task_const}
    \mathbf{g}(\{\mathbf{o}^k_i\}, \mathbf{h}_i, \mathbf{q}_i, \{\mathbf{b}^l_i\}) \leq 0,
\end{equation}
where $k \in [1,..K]$ is the index to an object, $l \in [1,..L]$ is the index to a body, and $\mathbf{o}^k_i \in O^k, \mathbf{h}_i \in H, \mathbf{q}_i \in Q, \text{ and } \mathbf{b}^l_i \in B^l$ are configuration vectors. With the explicit form we \micha{formulate} \textit{task constraints} \micha{ that describe} 
\theo{manipulation actions, \eg grasping and pushing actions---$\mathbf{g}(\mathbf{o}^k_i,  \mathbf{b}^l_i)$, and} relations in task space between objects and bodies, \eg collision between objects---$\mathbf{g}(\mathbf{o}^k_i,\mathbf{o}^{k'}_i)$. 
With \textit{task constraints} we also enforce \theo{kinematic and dynamic functions of multi-bodied systems, \eg forward/inverse kinematics of the human $\mathbf{g}(\mathbf{h}_i, \{\mathbf{b}^l_i\})$ and the robot $\mathbf{g}(\mathbf{q}_i, \{\mathbf{b}^l_i\})$.} 
\theo{\textit{Task constraints} are also used to represent human impairments, \eg \theo{ use of cane for support, limited joint ranges at the knee,} \etc.}
Each keyframe can have many \textit{task constraints}
that together specify \theo{critical points of the sequence (see~\cref{sec:intro}) or} mode switches (\textit{a.k.a.} transition manifolds) in MMMP or TAMP terminology~\cite{hauser2010multi} (\textit{a.k.a.} jumps in hybrid dynamical systems). 

\subsubsection*{\underline{Constancy constraints}} We use \textit{constancy constraints} to capture dependencies between keyframes, \theo{such as fixed pose between a hand and a bottle in three keyframes (i) grasping a bottle, (ii) holding it to pour into a mug and (iii) place it on the table, \ie \textit{constancy constraints} describe which motion dimensions between different keyframes remain constant.} 

Generally, a system's evolution within a mode \theo{(\textit{a.k.a.} system dynamics or flow in hybrid dynamical systems)} can be described using \textit{running constraints}~\cite{toussaint2022sequence}.  
\theo{
Their discrete form is $ f(\mathbf{x}_i, \mathbf{x}_{i+1}, \mathbf{u}_{i}) \leq 0$,} with $\mathbf{u}_{i}$ representing the control input. 
In case certain dimensions of a linear sequence remain constant \theo{(\eg stable grasps, contacts, \etc)} or bounded \theo{(\eg in-hand motions; an object sliding in the hand)}, these can be simplified to \textit{constancy constraints} $ \bar{f}(\mathbf{x}_i, \mathbf{x}_{i+1}) \leq 0$ that encode temporal indifferences between consecutive in time keyframes~\cite{toussaint2022sequence}.
Yet, when considering tree-like sequences \theo{(many possible concurrent futures)}, there is also the need to capture dependencies that occur between different keyframes of the same time-slice. Thus, we define \textit{constancy constraints} as
\vspace{-2mm}
\begin{equation}
 \label{eq:constancy_const}
 \mathbf{\bar{f}}(\mathbf{x}_i, \mathbf{x}_{j}) \leq 0,
\end{equation}
where $\mathbf{\bar{f}}(\cdot):\mathcal{X} \times \mathcal{X} \xrightarrow{}\mathbb{R}^{\gamma_{\mathbf{\bar{f}}}} $ with $\gamma_{\mathbf{\bar{f}}} \in \mathbb{R}$ specify a pair-wise coupling \theo{between $i$ and $j$ indexes of any two keyframes ($i \neq j$) even of 
the same time-slice.}

\SetKwComment{Comment}{/* }{ */}
\setlength{\textfloatsep}{0pt}
\begin{algorithm}[t]
\begin{scriptsize}
\caption{From \textit{task} to \textit{constancy constraints}}\label{alg:task2constancy}
\LinesNumbered
\KwIn{$\mathcal{T}$ \Comment*[r]{Set of \textit{task constr.}}}
\ForEach{$i \in [1, ..., N]$}{
    $ j \gets par(i)$ \Comment*[r]{Get parent index}
    \tcp{For any object or any body}
    \ForEach{$\omega \in [o^1, ..., o^K] \lor [b^1, ..., b^L] $}{
      \tcp{Check if parent \& child have same \textit{task constr.}}
      \If{   $\mathbf{g}(\omega_i) == \mathbf{g}(\omega_j)$}{
        \vspace{1mm} remove $\mathbf{g}(\omega_j)$ from $\mathcal{T}$; \\
        \vspace{1mm} add $\mathbf{\bar{f}}(\omega_i, \omega_j)$ in $\mathcal{T}$;
      }
    \tcp{For any object or any body}
    \ForEach{$\beta \in  [o^1, ..., o^K] \lor [b^1, ..., b^L] $ }{
      \tcp{Check if relative \textit{task constr.} between objects and/or bodies is constant}
      \If{   $\mathbf{g}(\omega_i, \beta_i) == \mathbf{g}(\omega_j, \beta_j)$}{
        \vspace{1mm}remove $\mathbf{g}(\omega_j, \beta_j)$ from $\mathcal{T}$; \\
        \vspace{1mm} add $\mathbf{\bar{f}}(\omega_i, \beta_i, \omega_j, \beta_j)$ in $\mathcal{T}$;
      }
    }
    }
}
\KwRet{$\mathcal{T}$ \Comment*[r]{Updated \textit{constr.} set}}
\end{scriptsize}
\end{algorithm}

\subsubsection*{\underline{Sequential invariance}}
\theo{Consider a keyframe where a human holds a bottle with the right hand, this keyframe can be followed by a keyframe where the human places the bottle on the table while still holding it with the right hand, but not by a keyframe where the human holds the same right hand a glass.}
In such a case, some dimensions of \theo{$ \mathcal{X}$ (see~\cref{par:influence_inter}) between adjacent\footnote{Adjacent keyframes are specified via parent-child relationship.} keyframes remain the same---invariant---\eg the distance between the right hand and the bottle (holding).}
Hence, such invariant dimensions can be enforced with \textit{constancy constraints} based on~\cref{alg:task2constancy}. 
 
\subsubsection*{\underline{Minimal description}}
We ensure that \textit{task} and \textit{constancy constraints} are described using a minimal set of coordinates, \theo{\eg an object placed upright on the table requires a 1-dof height and a 1-dof inclination constraint, and not a 6-dof pose constraint.}
\micha{Further, constraints are modelled as inequalities to allow task coordinates to vary within some given bounds,}
\micha{\eg the placement of an object on a table may be anywhere within the extents of the table surface.}

By exploiting the two above properties, we attain the maximal null space for each keyframe and \theo{we link the keyframes' null spaces along the sequence (linear or tree) to consider favorably interdependencies between keyframes and holistically optimize  objectives across keyframes.
This enables our method to reach better minima for the whole sequence as shown in~\cref{subsec:sim_study}.}

\subsection{Tree-Consistent-Constraint Problem}
\label{subsec:TCC}

By assembling a sequence of \textit{task} and \textit{constancy constraints} as described above, we can specify the schedule of manipulation tasks 
(\textit{a.k.a.} skeleton~\cite{busch2018planning}) with which we can predict how the human might execute the sequential task. \theo{We treat the prediction of the sequential task as \textit{primary objective}, \eg will the human pick the bottle or the mug, and the assistance as \textit{secondary objective} (inspired by the "assist-as-needed” principle), \eg where should the robot push the bottle}. Based on this strict 
hierarchy of objectives, we realize the RO problem \theo{(predict and assist)} defined in~\eqref{eq:RO} 
as 
\vspace{-1mm}
\begin{IEEEeqnarray}{CCCCC}
    \IEEEyesnumber\label{eq:1_ours} 
    \IEEEyessubnumber \label{eq:2_ours}
    \underset{\mathbf{x}_{1:N}}{\text{lexmin}}  &  \{\mathbf{c}_1\left(\mathbf{x}_{1:N}\right), \mathbf{c}_2\left( \mathbf{x}_{1:N}\right) \}~~~\hfill\\
    ~\text{s.t.}~  
    \IEEEyessubnumber \label{eq:3_ours} 
    & \mathbf{\bar{f}} \left(\mathbf{x}_{1:N}, \mathbf{x'}_{1:N}\right) + \mathbf{c}_1\left(\mathbf{x}_{1:N}\right) = 0,~\hfill\\
    \IEEEyessubnumber \label{eq:4_ours} &
    \mathbf{g} \left(\mathbf{x}_{1:N}\right) + \mathbf{c}_1\left(\mathbf{x}_{1:N}\right) =  0,~\hfill 
\end{IEEEeqnarray}
which is a lexicographic multi-objective problem~\cite{escande2014hierarchical}, \theo{\ie a problem with strict priority of objectives.}
$\mathbf{\bar{f}}(\cdot)$ and $\mathbf{g}(\cdot)$ can be stacked together as $\mathbf{T}_\mathcal{X}(\cdot)$ (see \cref{sec:prob_state} with $\gamma = d_{\mathbf{\bar{f}}} + d_{\mathbf{g}}$).
The cost term $\mathbf{c}_1(\cdot)$ is the \textit{primary objective} that is minimized first and it enacts as the barrier function to replace the inequalities~\eqref{eq:constancy_const} and~\eqref{eq:task_const} with equalities~\eqref{eq:3_ours} and~\eqref{eq:4_ours}. Here, it is also used as an innate metric of motion (\eg velocity minimization) that is often used in motion generation~\cite{zucker2013chomp}, \theo{\eg  a human posture is minimally changed between a standing keyframe and the following keyframe where the bottle is grasped}. 
\theo{The cost term $\mathbf{c}_2(\cdot)$ includes \textit{secondary objectives} that are minimized} only after the $\mathbf{c}_1(\cdot)$ and is an expectation over all possible sequences (see eq. 9.7 in~\cite{kochenderfer2022algorithms}). Its metric functions depend on \theo{comfort, ergonomics, \etc, which} is implicitly influenced by the robot interventions. 
By solving~\eqref{eq:1_ours} we can obtain a sequence (linear or tree) of N configurations $\mathbf{x}_{1:N} \in \mathbb{R}^{N \times d_\mathbf{x}}$. This folds within; predictions of multiple futures of the human and objects states (see~\eqref{eq:pred-only}), and simultaneously a single sequence of optimal robot interventions for all possible futures  (see~\eqref{eq:RO}).

\subsubsection*{\underline{Solver}}

\theo{To solve~\eqref{eq:1_ours}, we treat it as a
non-linear least-squares problem\footnote{\theo{Non-linear least-squares is a commonly used unconstrained optimization formulation of IK and  estimation (\eg SLAM~\cite{factor_graphs_for_robot_perception}) problems in robotics.}} with hierarchy and we construct its feasible region using a barrier function based on displacement intervals~\cite{gienger2006exploiting}.}
By linearizing the non-linear constraint functions using a Taylor expansion, we can obtain the Jacobian\footnote{The \textit{constancy constraints} $ \mathbf{\bar{f}}(\cdot)$ make the Jacobian non-banded but sparse.} of the constraints $\mathbf{J}$, and we impose the strict hierarchy by projecting secondary objectives in the null space of the primary ones~\cite{slotine1991general}. 
Following this strategy the problem in~\eqref{eq:1_ours} can be efficiently solved with a series of Levenberg–Marquardt updates, based on
\begin{equation}
\hspace{-5mm}
\begin{aligned}
    \label{eq:Liegeois}
    &\Delta\mathbf{x}_{1:N} =    
    &\underbrace{ {\scriptsize\mathbf{J}^{\#} \left(\Delta\mathbf{T}_\mathcal{X}  + \nabla{ \mathbf{c}_{1}^T}\mathbf{J}^{\#}\right)}}_{\theo{\text{prediction term}}}  + \underbrace{ {\scriptsize\mathbf{A} \mathbf{P} } \nabla{ \mathbf{c}_{2}^T}}_{\theo{\text{assistance term}}},    \end{aligned} 
    \hspace{-4mm}
    \vspace{-2mm}
\end{equation}
where
\begin{equation}
    \label{eq:LMU}
    \mathbf{J}^\# = (\mathbf{J}^T \mathbf{W}_{\mathbf{x}_{1:N}} \mathbf{J}+ \lambda \mathbf{I})^{-1} \mathbf{J}^T \mathbf{W}_{\mathbf{x}_{1:N}},
\end{equation}
is the pseudoinverse Jacobian of 
\micha{all constraints of all keyframes,}
$\mathbf{W}_{\mathbf{x}_{1:N}} \in \mathbb{R}^{d_\mathbf{x} \times d_\mathbf{x}}$ is a weight matrix, 
$\mathbf{P} = (I - \mathbf{J}^\# \mathbf{J})\mathbf{W}^{-1}_{\mathbf{x}_{1:N}}$ is a projector on the null space of $\mathbf{J}$, and $\mathbf{A}$ is a selection matrix that filters  per dimension \theo{the interventions resulting from $\mathbf{c}_2(\cdot)$ (secondary objective in~\eqref{eq:1_ours})}. $\nabla{ \mathbf{c}_1(\mathbf{x}_{1:N})} \in \mathbb{R}^\gamma$ acts as the barrier function and  is the gradient of $\mathbf{c}_1(\cdot)$ \theo{(primary objective in~\eqref{eq:1_ours})} with respect to $\mathbf{x}_{1:N}$. Scalar $\lambda$ is a damping parameter that allows to switch between Gauss-Newton (small $\lambda$) and gradient descent updates (large $\lambda$).

In this way, we can guarantee that the prediction of all possible futures is \theo{attained first, which corresponds to the first focal point in~\cref{sec:prob_state}}. The assistance term \theo{(second focal point in~\cref{sec:prob_state})} lies in the null space of the (multiple) predictions, hence, the interventions are optimal for all possible futures concurrently (expectation over futures), which satisfy the RO problem stated in~\eqref{eq:RO}.

%% file: figures/methods/concept-seqManip-drawing.tex
\begingroup%
  \makeatletter%
  \providecommand\color[2][]{%
    \errmessage{(Inkscape) Color is used for the text in Inkscape, but the package 'color.sty' is not loaded}%
    \renewcommand\color[2][]{}%
  }%
  \providecommand\transparent[1]{%
    \errmessage{(Inkscape) Transparency is used (non-zero) for the text in Inkscape, but the package 'transparent.sty' is not loaded}%
    \renewcommand\transparent[1]{}%
  }%
  \providecommand\rotatebox[2]{#2}%
  \newcommand*\fsize{\dimexpr\f@size pt\relax}%
  \newcommand*\lineheight[1]{\fontsize{\fsize}{#1\fsize}\selectfont}%
  \ifx\svgwidth\undefined%
    \setlength{\unitlength}{1275.59055118bp}%
    \ifx\svgscale\undefined%
      \relax%
    \else%
      \setlength{\unitlength}{\unitlength * \real{\svgscale}}%
    \fi%
  \else%
    \setlength{\unitlength}{\svgwidth}%
  \fi%
  \global\let\svgwidth\undefined%
  \global\let\svgscale\undefined%
  \makeatother%
  \begin{picture}(1,0.33333333)%
    \lineheight{1}%
    \setlength\tabcolsep{0pt}%
    \put(0,0){\includegraphics[width=\unitlength,page=1]{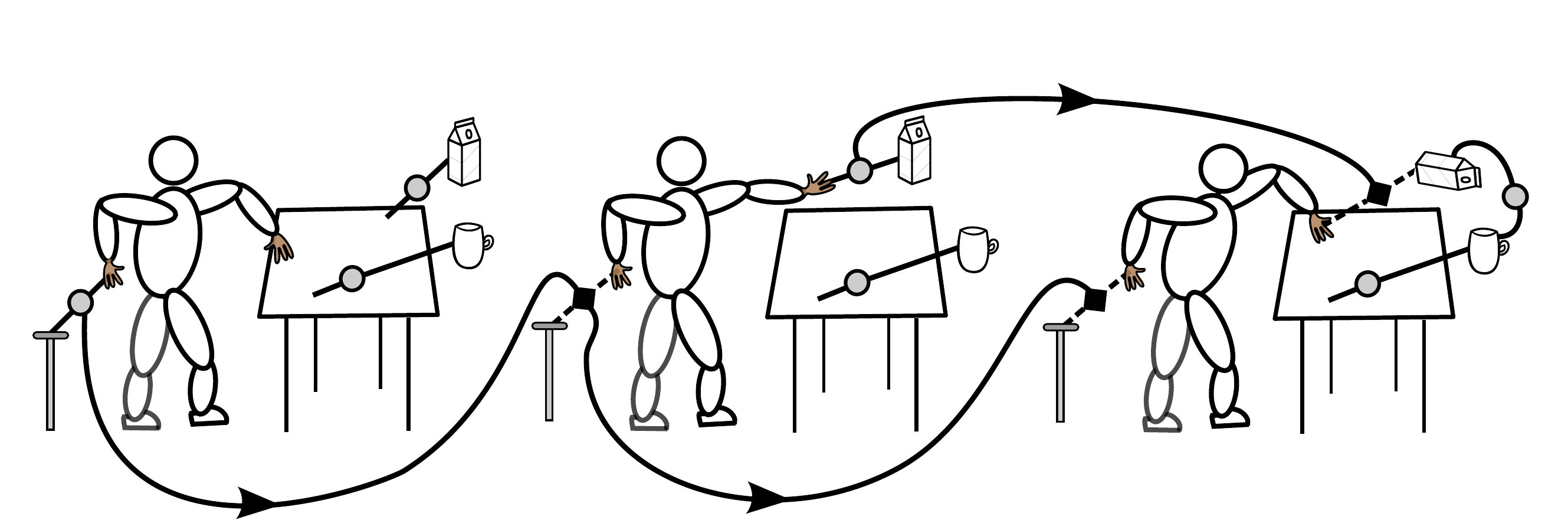}}%
    \put(0.15786514,0.29819078){\color[rgb]{0,0,0}\makebox(0,0)[lt]{\lineheight{1.25}\smash{\begin{tabular}[t]{l}\footnotesize{\textit{task constraints}}\end{tabular}}}}%
    \put(0,0){\includegraphics[width=\unitlength,page=2]{figures/methods/concept-seqManip-drawing-1.pdf}}%
    \put(0.65195001,0.29925669){\color[rgb]{0,0,0}\makebox(0,0)[lt]{\lineheight{1.25}\smash{\begin{tabular}[t]{l}\footnotesize{\textit{constancy constraints}}\end{tabular}}}}%
  \end{picture}%
\endgroup%

%% file: sections/Evaluations.tex
In this section, we first study the importance of the proposed constraint formulation (see~\cref{subsec:constraints}) in terms of interventions by performing an ablation study in two scenarios. These scenarios also demonstrate the 
ability of the method to consider both linear and tree sequences. Further, we show the our method's capability to personalize the predicted human action based on various factors such as impairments, injuries, strength, \etc. 
Finally we realize the assistive capabilities of the method in a table-top scenario with a bimanual robot that provides personalized support to improve the ergonomic posture of a human while preparing a beverage. See the attached video for simulations and experiments of the above described capabilities.

\subsubsection*{Implementation setup}  
\theo{
Our method generate keyframes sequences 
for a beverage pouring task, omitting continuous trajectories.} 
All evaluations are conducted on a 64-bit Intel Quad-Core i5 3.80 GHz computer with 16GB RAM. The concepts are implemented in the \texttt{Rcs} C++ library\footnote{\href{url}{https://github.com/HRI-EU/Rcs}}.

\begin{figure}[t]
    \centering
    \subfloat[Linear sequence \theo{in simulation}. From left to right, keyframes; (2) pick a bottle, (3) pour into glass, (4) place bottle on table, and (5) pick glass, while using the right hand only, as the left holds the cane for support. Intervention: relocate glass before keyframe (3).]{\includegraphics[width=0.95\linewidth, frame]{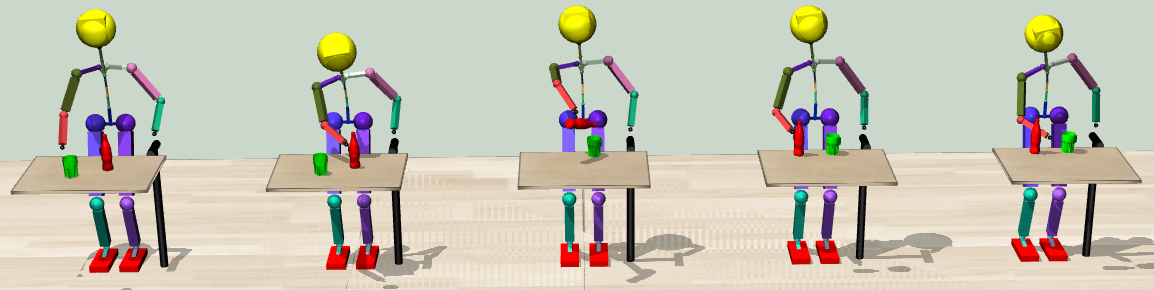}}~\\
    \def\svgwidth{0.95\linewidth}\input{figures/simulation_results/Label.tex} \vspace{-4mm}\\ 
    \newcommand\fw{0.25\linewidth}
    \subfloat[\centering REBA.]{\includegraphics[width=\fw]{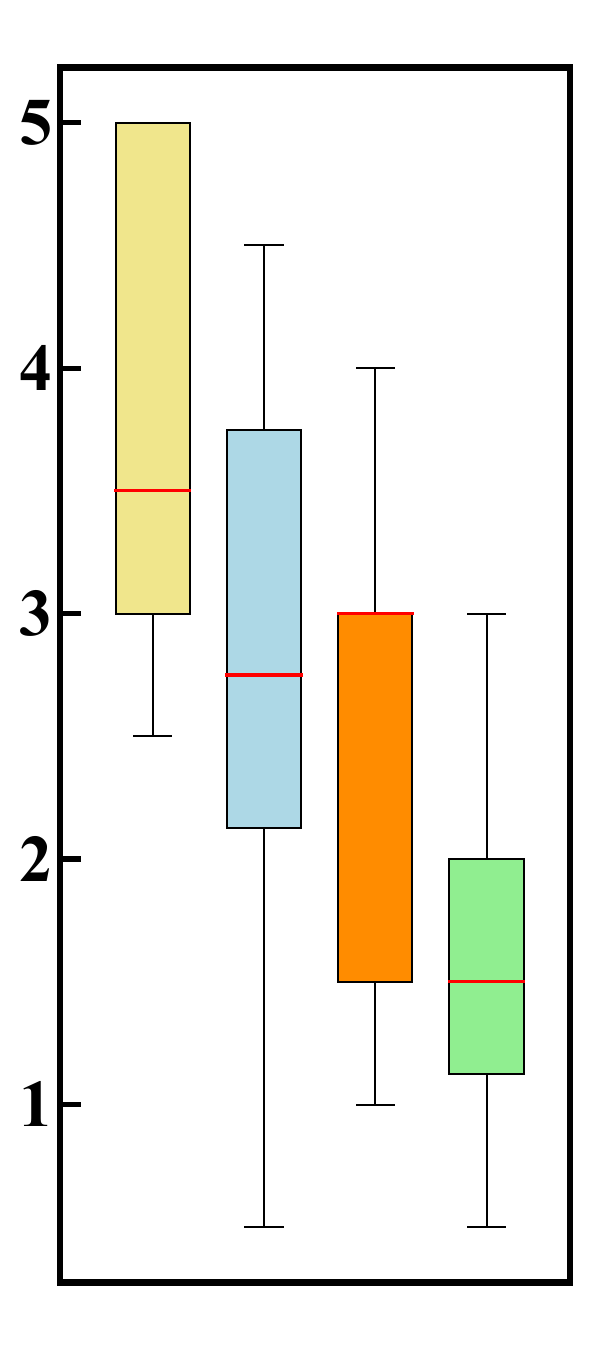}}~
    \subfloat[\centering Static effort (\unit{(N \cdot m)^2}).]{\includegraphics[width=\fw]{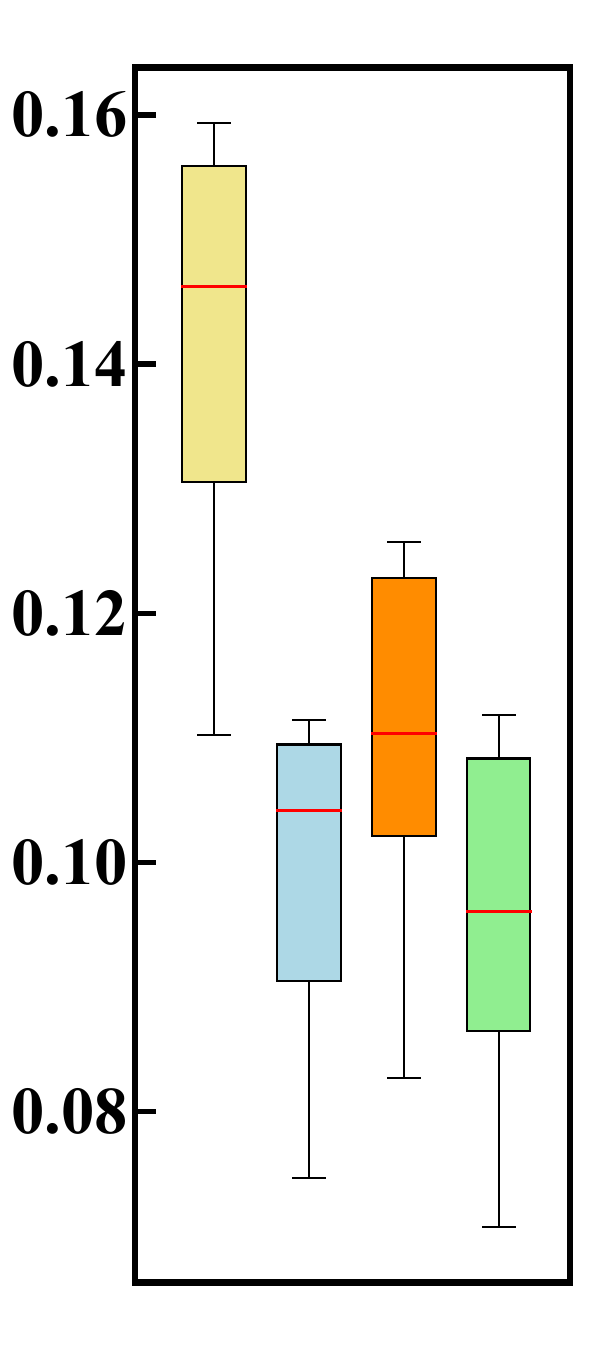}}~
    \subfloat[\centering Spine loading (\unit{(N \cdot m)^2}).]{\includegraphics[width=\fw]{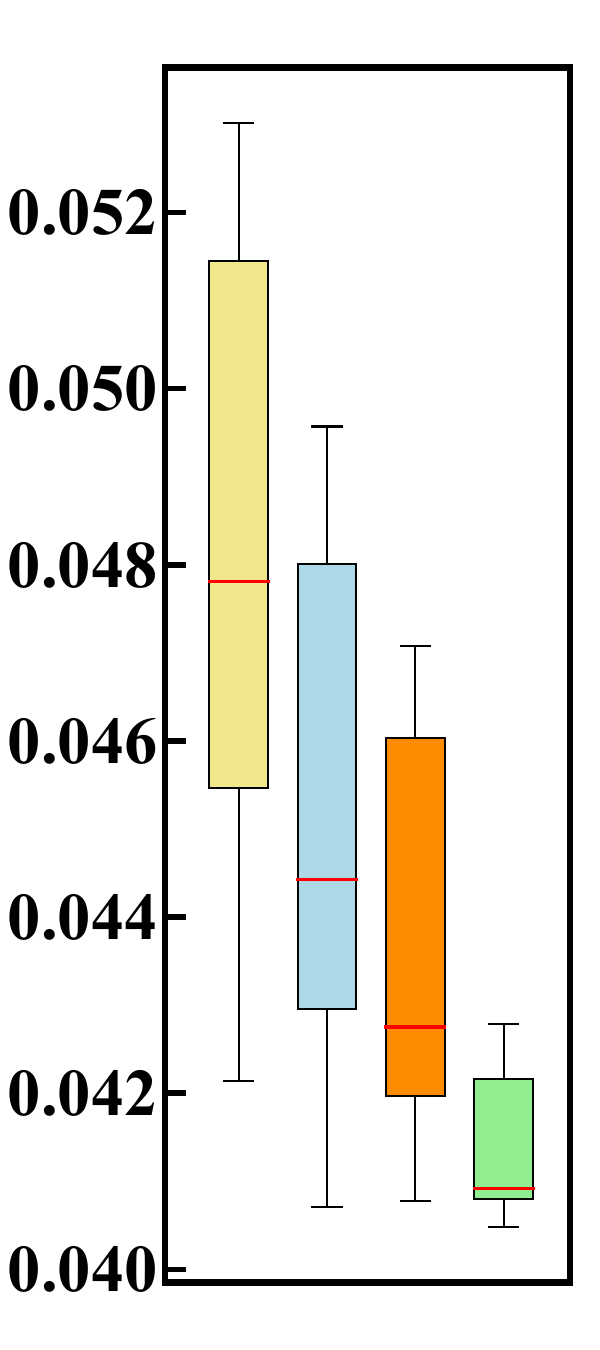}}~
    \caption{Average metrics \theo{on keyframes 3-5 (post-intervention)} for the four methods on the task shown in (a).} 
    \label{fig:resultplt_linear}
    \vspace{2mm}
\end{figure}

\begin{figure}[t]
    \centering
    \subfloat[\theo{Tree sequence in simulation}. Pick the mug with right or left hand using power-grasp and hand it over to the free hand to pinch-grasp the handle. Intervention: relocate glass to a common pose for both keyframes 2 and 4.]{\includegraphics[width=0.8\linewidth, frame]{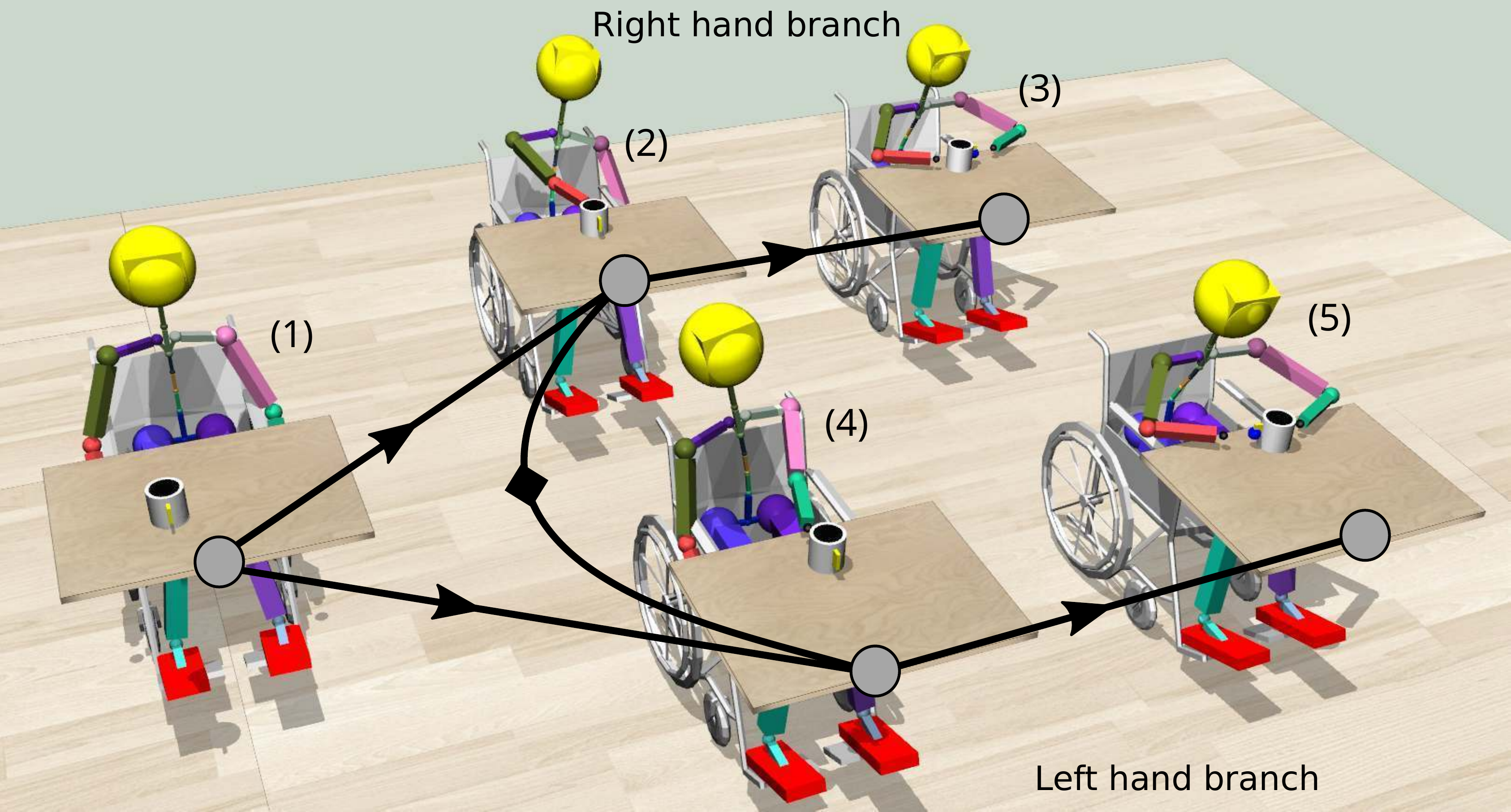}}~\\
    \def\svgwidth{0.95\linewidth}\input{figures/simulation_results/Label.tex} \vspace{-4mm}\\ 
    \newcommand\fw{0.25\linewidth}
    \subfloat[\centering REBA.]{\includegraphics[width=\fw]{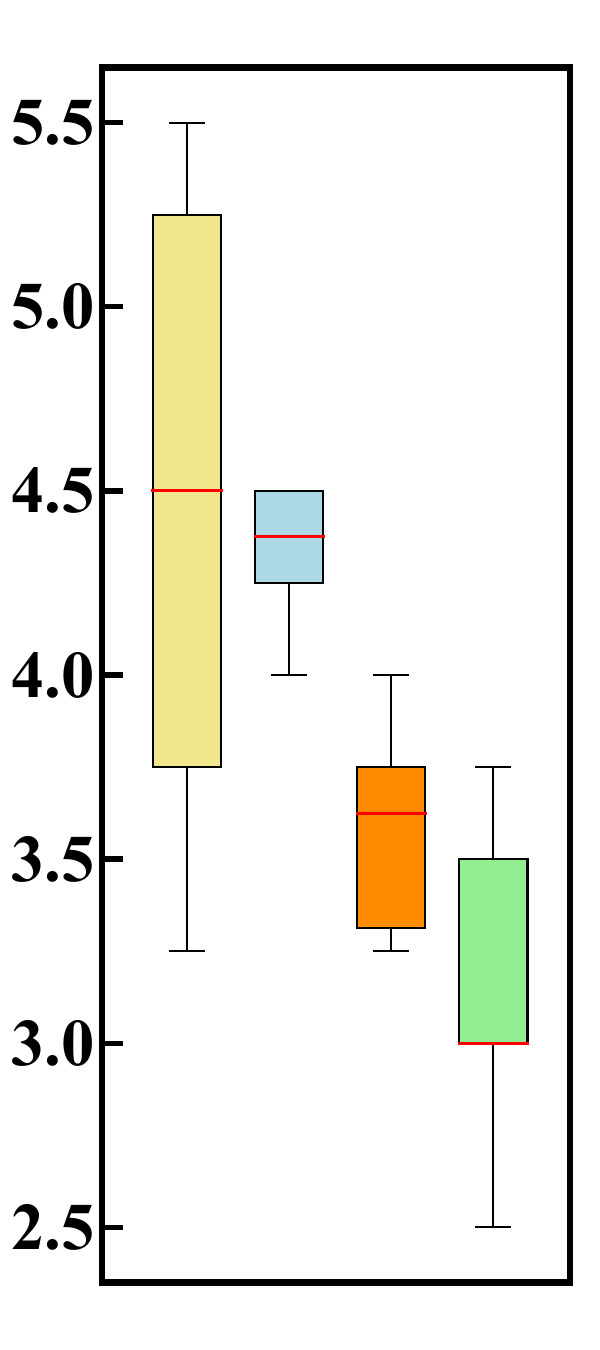}}~
    \subfloat[ \centering Spine loading (\unit{(N \cdot m)^2}).]{\includegraphics[width=\fw]{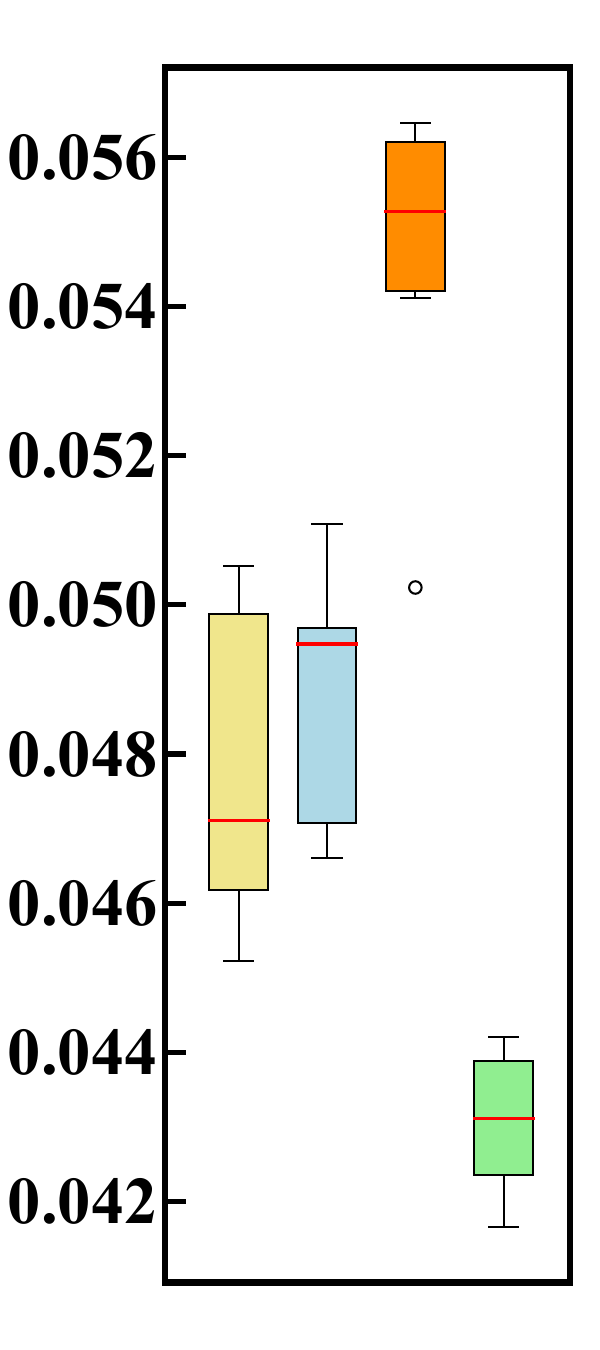}}~
    \subfloat[\centering  Joint limit \newline violations (rad).]{\includegraphics[width=\fw]{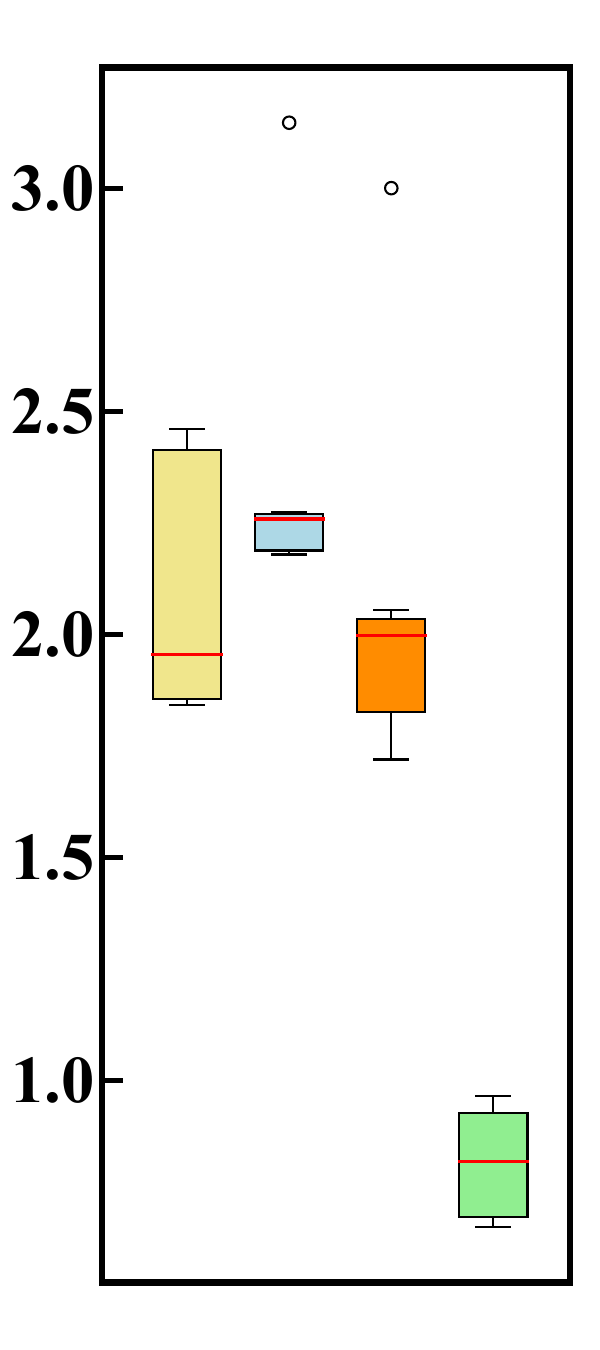}}~
    \caption{Average metrics \theo{on keyframes 2-5 (post-intervention)} for the four methods on the task shown in (a). } 
    \label{fig:resultplt_tree}
    \vspace{2mm}
\end{figure}

\subsection{\theo{Comparative studies in simulation}}
\label{subsec:sim_study}
Here, we compare \theo{three methods against ours to investigate how different formulations of constraints $ \mathbf{T}_\mathcal{X}(\cdot) \leq 0 $ influence optimality when using the solver in~\eqref{eq:Liegeois}}:
\begin{itemize}
    \item \textit{\theo{(i) No intervention}}: Only prediction of the human's future actions is realized without any intervention.  Essentially, we solve the problem defined in~\eqref{eq:pred-only}.
    \item \textit{Heuristic-based}: These method optimizes the configuration of each keyframe separately without considering transitions from one keyframe to another. Then, for all free dimensions that are common between linked keyframes, strategy \textit{\theo{(ii) First}} applies the configuration of the first keyframe to all consecutive ones. Strategy \textit{ \theo{(iii) Average}} computes the average configuration of the keyframes and applies them to all linked keyframes. 
    \item \theo{\textit{(iv) Coupled}: Our approach that solves~\eqref{eq:RO} and~\eqref{eq:1_ours} to concurrently predict the human's future actions and find the optimal intervention across linked keyframes.}  
\end{itemize}

\noindent For all methods we use the following optimization criteria (across keyframes) that are:
\begin{itemize}
    \item \textit{Human posture}: A sum-of-square (SOS) minimization of all human joints to a preferred configuration (posture). This term penalizes joint limit violations and the preferred posture has REBA score equal to 1.
    \item \textit{Static effort}: A SOS minimization of all human static joint torques for a given load at the right hand of the human. This incentivizes to avoid stretching the arm.
    \item \textit{Spine loading}: A SOS minimization of the spine joint torques with respect to a given load at a head of the human. This incentivizes to avoid bending.  
\end{itemize}

\noindent The scenarios are; first, a linear sequential manipulation task shown in~\cref{fig:resultplt_linear}(a), where a human uses a cane for support. \theo{The glass location in keyframes 3 (pouring) and 5 (picking) are the same, hence linked with \textit{constancy constraint}.}
Yet, due to different orientation and height of the pouring and the picking actions, the optimal human postures and glass poses in keyframes 3 and 5 differ. The robot's intervention takes place between keyframe 2 and 3. The location of the glass (green) is modified so that the above criteria are optimized. \vspace{-1mm}

Second, a tree sequential manipulation task is
shown in~\cref{fig:resultplt_tree}(a), where a human is on a wheelchair and either power-grasps the mug with the right hand to handed it over to the left for a pinch-grasp from the handle, or \theo{power-grasps the mug with the left to pinch-grasp it from the handle with the right. The task is motivated from scenarios where; (i) the grasping actions is ambiguous (power-grasp with left or right hand) and (ii) the power-grasp orientation of the first determines the feasibility of the pinch-grasp (second action).}
The robot's intervention takes place between keyframe 1 and both 2 and 3. The location of the mug (gray) is modified so that the objectives of both futures (right and left hand branches) are optimized, \theo{\ie both branches can power-grasp the mug with orientations that allow pinch-grasp from the handle in the next action.} \vspace{-1mm}

The metrics uses to evaluate the methods are averaged across keyframes, (i) REBA score where we subtract 1 as it is the minimum, (ii) \textit{Static effort}, (iii) \textit{Spine loading}, and (iv) joint limit violations. We sample 10 different poses (position and orientation) of the human (with the cane) and of the wheelchair. We report mean and standard deviations with the bar plots shown in ~\cref{fig:resultplt_linear} and ~\cref{fig:resultplt_tree} for each scenario.

As it can be observed, ours (\textit{Coupled}) outperforms the \textit{no intervention} method, \theo{which demonstrates that the interventions of our method
improve the human state. The \textit{Coupled} method also performs preferably to the two heuristic methods}, which illustrates that having the constraint structure, described in~\cref{subsec:constraints}, forms a null-space that can be exploited to reach better optima by our solver (see~\cref{subsec:TCC}). The optimality of two heuristic methods depends on the scenario and the metric, which motivates the development of a consistent method, like the \textit{Coupled} (ours).

\subsection{Simulation based case studies}

\subsubsection{Utility of \textit{constancy constraints} within a time-slice}
 
\begin{figure}[t]
    \centering
    \includegraphics[width=0.8\linewidth, frame]{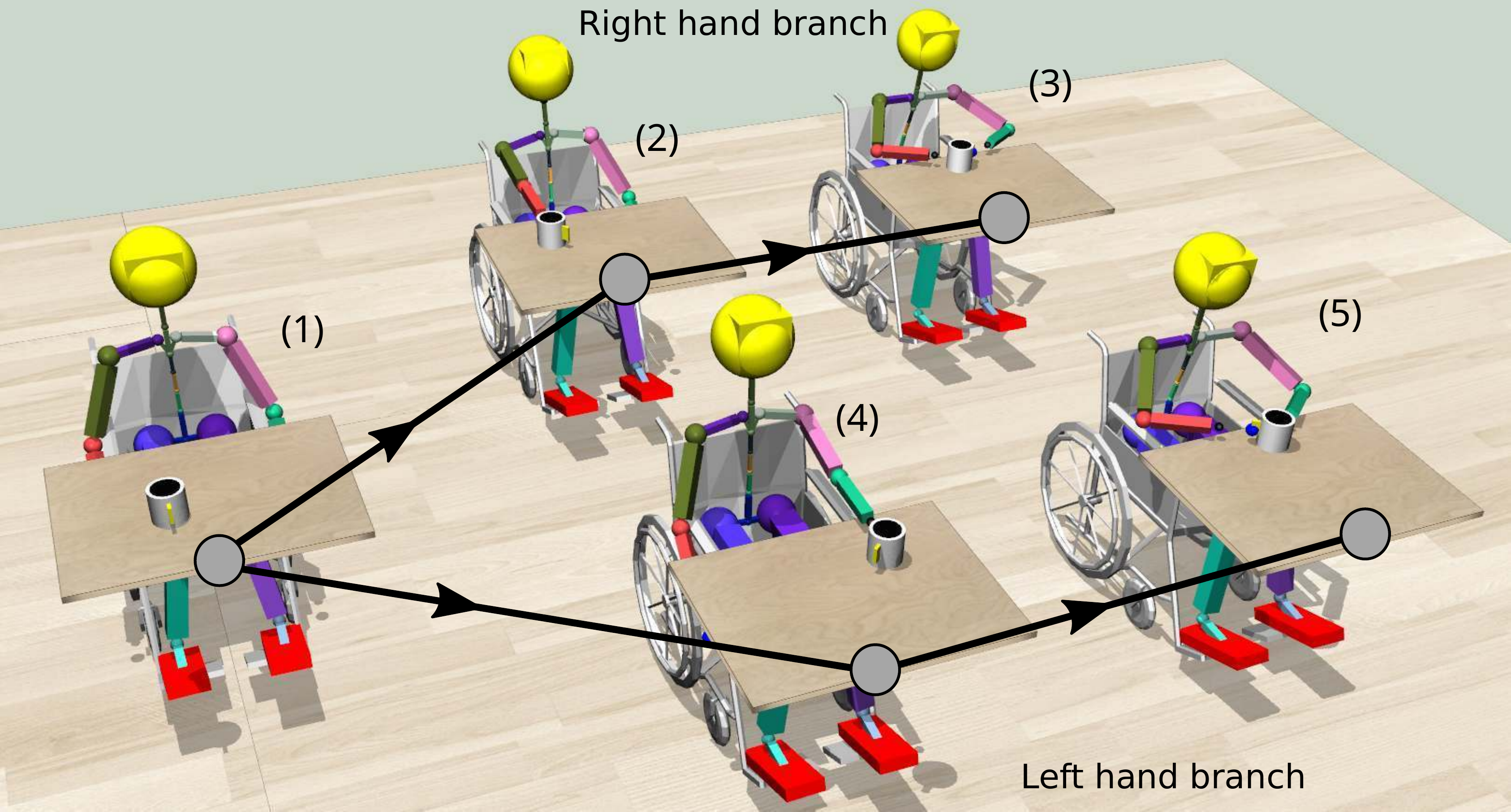}
    \vspace{-2mm}
    \caption{Divergent tree sequence, where the right and the left hand branched are independent. Intervention: relocate glass for the right and the left branch separately. See keyframes 2 and 4.  }
    \label{fig:decoupled_tree}
    \vspace{1mm}
\end{figure}

 \cref{fig:decoupled_tree} shows a tree sequence with two possible futures along with the graph \theo{connectivity} of the linked keyframes. \theo{This indicates that the two independent branches of the graph (right and left hand) have different optimal poses of the mug. }
 If the wrong future is assumed, then the intervention is not optimal. In contrast, the coupled method (see~\cref{fig:resultplt_tree}(a)) finds a common optimal pose for the mug for both futures (right or left hand branches). \theo{This is enabled by the
 \textit{constancy constraint} (see the diamond) that links 
 keyframes 2 and 4}. 
 
\subsubsection{Personalized predictions}

\begin{figure}[t]
    \centering
    \subfloat[Nominal prediction with a subject requiring a cane with the left hand for stabilization. ] 
    {\includegraphics[width=0.75\linewidth, frame]{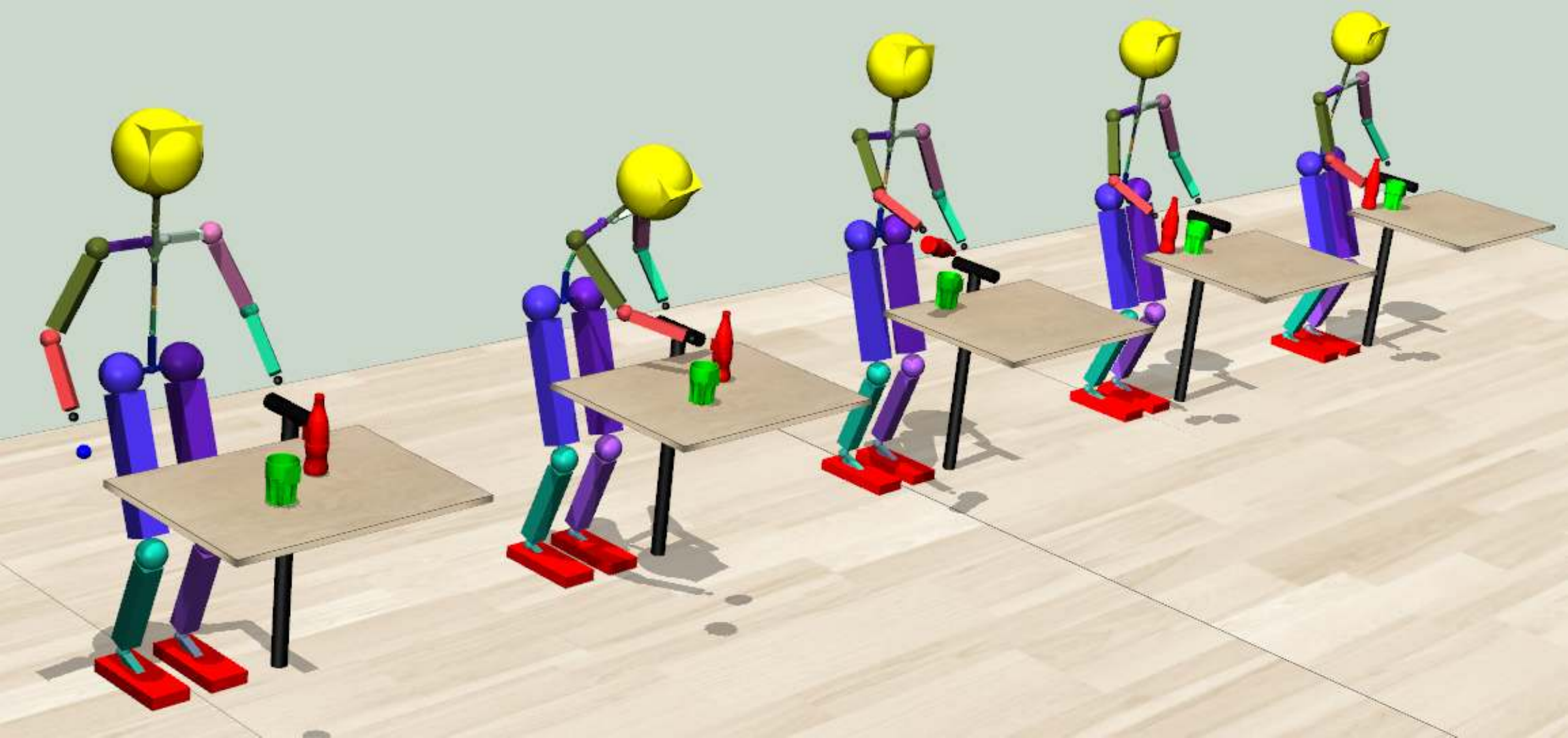}}~\\
    \subfloat[Impaired subject same as (a) and with immobilized (broken) right elbow.] 
    {\includegraphics[width=0.75\linewidth, frame]{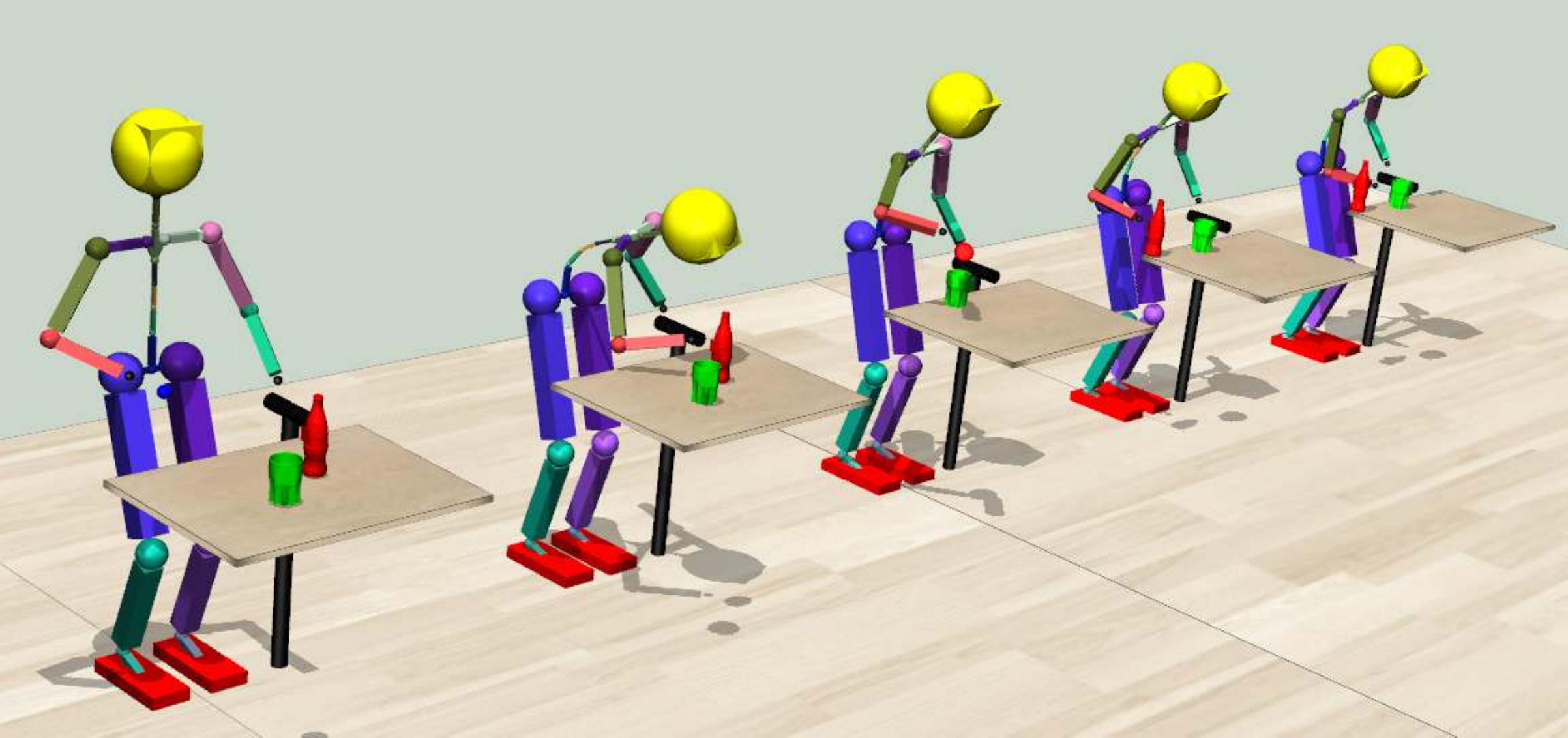}}~\\
    \subfloat[Impaired subject same as (a) and requiring a large stabilizing force on the cane with the left hand.] 
    {\includegraphics[width=0.75\linewidth, frame]{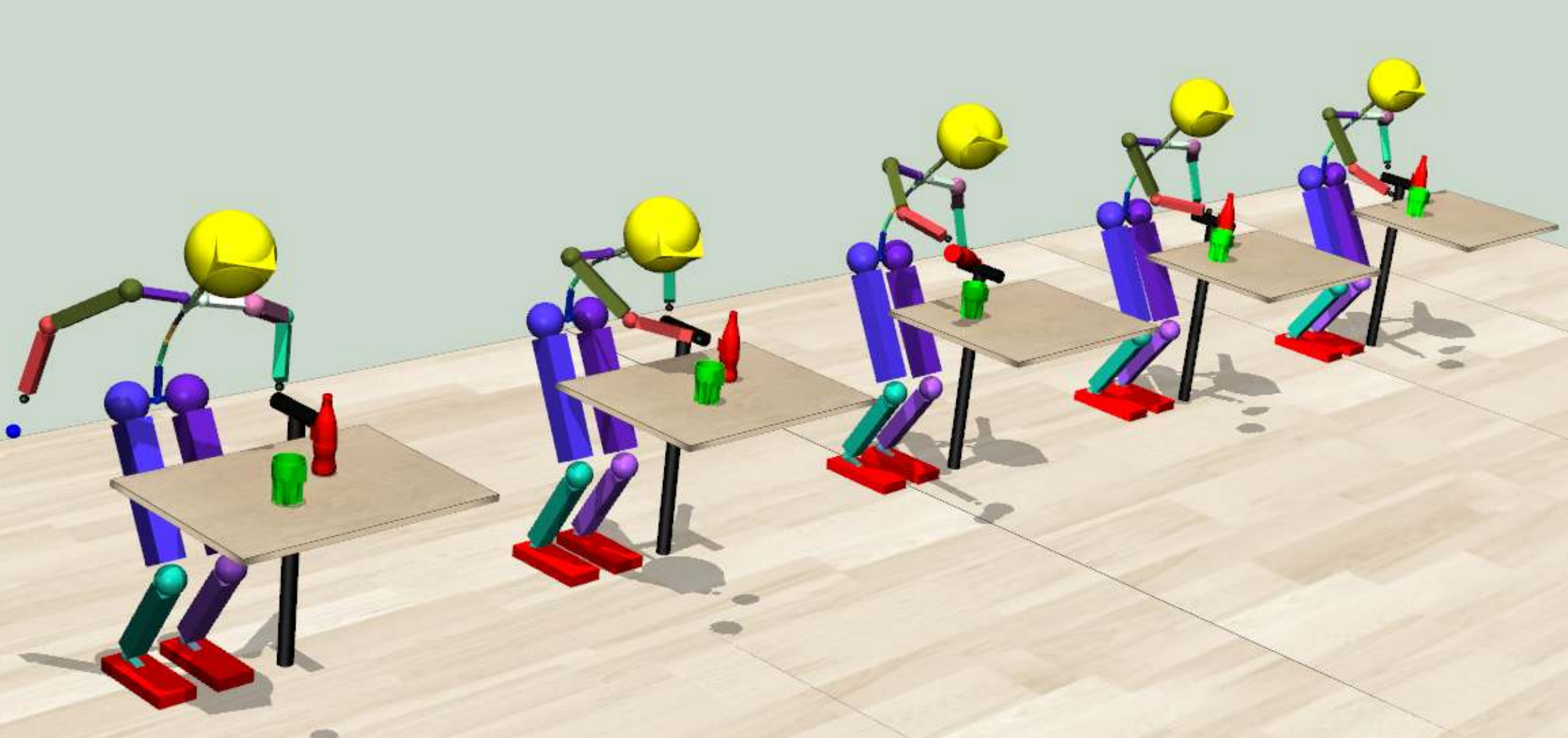}}~\\
    \caption{ Simulation of predicted postures based on human impairments. } 
    \label{fig:injury_based_prediction}
    \vspace{2mm}
\end{figure}

\cref{fig:injury_based_prediction}(a) shows a nominal prediction where the robot's intervention takes place between keyframe 2 and 3.
\theo{\cref{fig:injury_based_prediction}(b) shows the same sequence, however with a different human impairment; an immobilized (broken) right elbow.}
This impairment leads to a strong bending motion of the upper body to reach the bottle and the glass to compensate for the immobilized elbow. \cref{fig:injury_based_prediction}(c) is a result of an \theo{analytically} modelled impairment (\eg weak left leg) that makes the human lean strongly onto the cane.

In both cases, the impairment \theo{consideration} leads to a different support intervention which positions the glass in keyframe 3 at a different location to the one in 
\cref{fig:injury_based_prediction}(a).
These cases demonstrate the capability of our method to perform personalised predictions and interventions. \vspace{-1mm}

\begin{figure*}[t]
    \centering
    \subfloat[Upright posture with extended arm.]{\includegraphics[width=0.85\linewidth, frame]{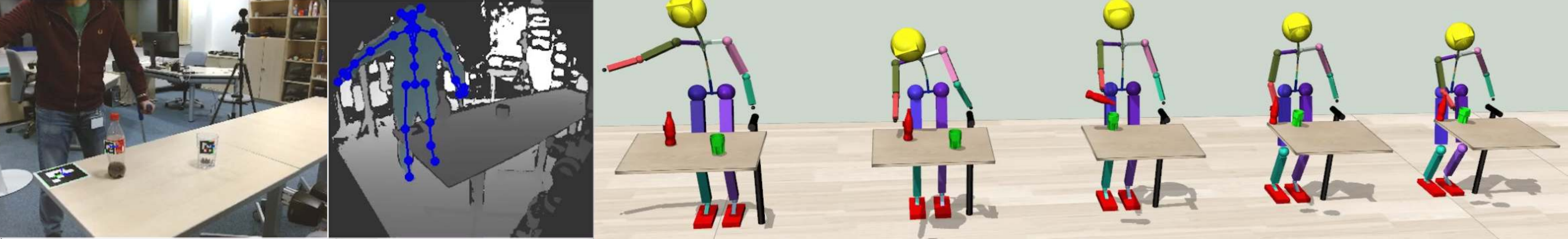}}~\\
    \subfloat[Back injury.]{\includegraphics[width=0.85\linewidth, frame]{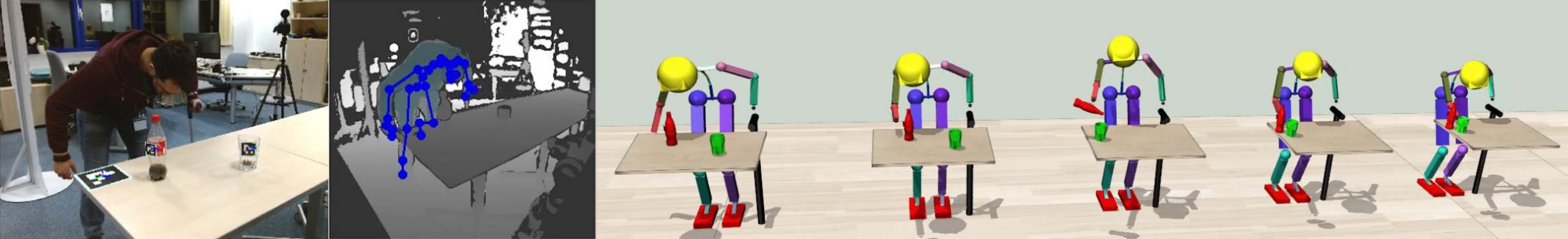}}~\\
    \caption{Interactive system that utilizes 5 Hz sensor updates to update prediction and interventions on-the-fly. } 
    \label{fig:azure_based_prediction}
\end{figure*}

\subsection{Robot experiments}

\cref{fig:azure_based_prediction} illustrates the system operating interactively in a dynamically sensed environment. The location of the table, bottle, and glass are detected using fiducial markers. The posture of the human is determined by a 3d posture tracking system (Azure Kinect DK) \theo{with 5Hz. This enables us to compute the REBA score of the sensed human. The sensed human posture is mapped onto keyframe 1 and it is assumed to be the preferred pose. The predicted human postures (rest keyframes) are updated on-the-fly by solving the TCC problem. One iteration of the optimization takes about $50-70$~msec, leading to convergence in 1-2 seconds.}
In~\cref{fig:azure_based_prediction}(a), the human is in an upright posture, while in~\cref{fig:azure_based_prediction}(b) the human posture 
reflects a back injury (leaning forward). As in the simulations above, the predicted keyframes 
show personalised predictions and the effect of the robot's  intervention; to relocate the glass between keyframe 2 and 3.

\begin{figure*}[t]
    \centering
    \subfloat[Unassisted. REBA score during pouring and picking the glass varies between 4 and 5.]{\includegraphics[width=0.85\linewidth, frame]{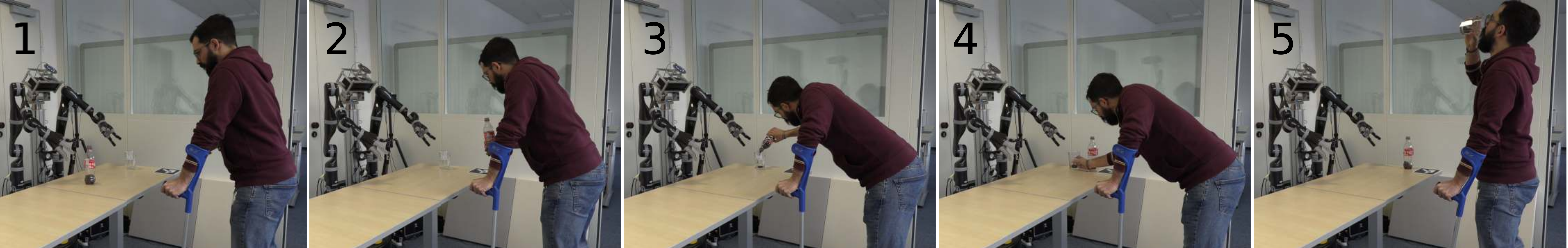}}~\\ \vspace{-1mm}
    \subfloat[Assisted. REBA score during pouring and picking the glass varies between 2 and 3.]{\includegraphics[width=0.85\linewidth, frame]{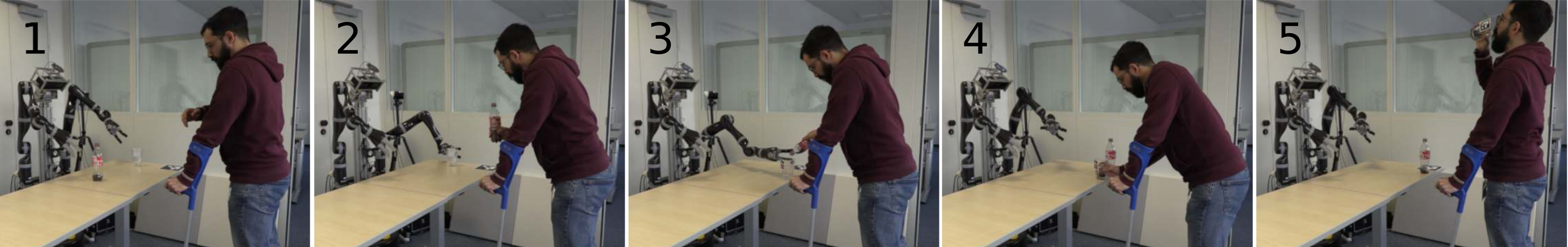}}~\\
    \caption{Keyframes of the robot-assisted pour-bottle-into-glass task. By comparing keyframes 3 and 4 for the two cases, we can observe the effect that the robot interventions have in the human posture. } 
    \label{fig:experiments}
    \vspace{-5mm}
\end{figure*}

Further, we validate the utility of the proposed approach to indeed improve the posture of the human. As shown in~\cref{fig:experiments}, based on the sensor-based predictions and intervention, a robot assists a human during the execution of a pour-bottle-into-glass task by pushing the glass to the preferable pose. In contrast to the assisted case, in~\cref{fig:experiments}(a) the human needs to bend forward extensively to complete the task. Further robot support intervention cases can be observed in the video. \vspace{-1mm}

%% file: figures/simulation_results/Label.tex
\begingroup%
  \makeatletter%
  \providecommand\color[2][]{%
    \errmessage{(Inkscape) Color is used for the text in Inkscape, but the package 'color.sty' is not loaded}%
    \renewcommand\color[2][]{}%
  }%
  \providecommand\transparent[1]{%
    \errmessage{(Inkscape) Transparency is used (non-zero) for the text in Inkscape, but the package 'transparent.sty' is not loaded}%
    \renewcommand\transparent[1]{}%
  }%
  \providecommand\rotatebox[2]{#2}%
  \newcommand*\fsize{\dimexpr\f@size pt\relax}%
  \newcommand*\lineheight[1]{\fontsize{\fsize}{#1\fsize}\selectfont}%
  \ifx\svgwidth\undefined%
    \setlength{\unitlength}{870bp}%
    \ifx\svgscale\undefined%
      \relax%
    \else%
      \setlength{\unitlength}{\unitlength * \real{\svgscale}}%
    \fi%
  \else%
    \setlength{\unitlength}{\svgwidth}%
  \fi%
  \global\let\svgwidth\undefined%
  \global\let\svgscale\undefined%
  \makeatother%
  \begin{picture}(1,0.08045977)%
    \lineheight{1}%
    \setlength\tabcolsep{0pt}%
    \put(0,0){\includegraphics[width=\unitlength,page=1]{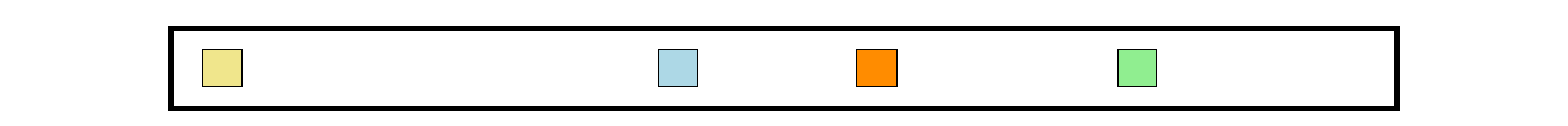}}%
    \put(0.17,0.02682635){\color[rgb]{0,0,0}\makebox(0,0)[lt]{\lineheight{1.25}\smash{\begin{tabular}[t]{l}\footnotesize{No intervention}\end{tabular}}}}%
    \put(0.46,0.02682635){\color[rgb]{0,0,0}\makebox(0,0)[lt]{\lineheight{1.25}\smash{\begin{tabular}[t]{l} \footnotesize{First}\end{tabular}}}}%
    \put(0.58,0.02682635){\color[rgb]{0,0,0}\makebox(0,0)[lt]{\lineheight{1.25}\smash{\begin{tabular}[t]{l}\footnotesize{Average}\end{tabular}}}}%
    \put(0.75,0.02682635){\color[rgb]{0,0,0}\makebox(0,0)[lt]{\lineheight{1.25}\smash{\begin{tabular}[t]{l}\footnotesize{Coupled}\end{tabular}}}}%
  \end{picture}%
\endgroup%

%% file: sections/conclusion.tex
We have proposed a novel concept to support impaired users in daily physical object manipulation tasks with a robot. Our concept is based on a prediction model that uniquely casts the user’s sequential behavior as well as a robot support intervention into a hierarchical multi-objective optimization problem. This allows to compute an optimal robot support intervention that rearranges objects in the scenario so that the impaired human is optimally supported. We have shown in simulation studies that our concept is effective, both in terms of the human's comfort metric and in computational efficiency to enable on-the-fly updates of the predictions and the interventions. Real-world experiments highlight the ability to provide support in dynamically changing situations. 

Future work will focus on relaxing the assumptions related to the sequence of the task, \theo{on data-driven models of human impairments and on utilization of Augmented Reality to enhance the interpretability of the assistive robot~\cite{wang2023explainable}.}
\vspace{-2mm}